\PassOptionsToPackage{dvipsnames}{xcolor}

\documentclass{article}
\usepackage[utf8]{inputenc} % allow utf-8 input
\usepackage[T1]{fontenc}    % use 8-bit T1 fonts
\usepackage{hyperref}       % hyperlinks
\usepackage{url}            % simple URL typesetting
\usepackage{booktabs}       % professional-quality tables
\usepackage{amsfonts}       % blackboard math symbols
\usepackage{nicefrac}       % compact symbols for 1/2, etc.
\usepackage{microtype}      % microtypography
\usepackage{lipsum}		% Can be removed after putting your text content
\usepackage{graphicx}
\usepackage{float}
\usepackage{natbib}
\usepackage{doi}
\usepackage{xcolor}
\usepackage{lineno}
\usepackage{amsmath}
\usepackage{xcolor}
\usepackage{arxiv}
\usepackage{url}
\newcommand\myshade{85}
\colorlet{mylinkcolor}{RoyalBlue}
\colorlet{mycitecolor}{violet}
\colorlet{myurlcolor}{YellowOrange}

\hypersetup{
  linkcolor  = mylinkcolor!\myshade!black,
  citecolor  = mycitecolor!\myshade!black,
  urlcolor   = myurlcolor!\myshade!black,
  colorlinks = true,
}

\usepackage{times}
\usepackage{latexsym}

\usepackage[T1]{fontenc}
\usepackage[utf8]{inputenc}

\usepackage{microtype}

\usepackage{booktabs}
\usepackage{graphicx}
\usepackage{makecell}
\usepackage{multirow}
\usepackage{longtable}
\usepackage{CJKutf8}
\usepackage{float}

\usepackage{tikz}
\usetikzlibrary{shapes.geometric}

\newcommand\score[2]{%
  \pgfmathsetmacro\pgfxa{#1 + 1}%
  \tikzstyle{scorestars}=[star, star points=5, star point ratio=2.25, draw, inner sep=1.3pt, anchor=outer point 3]%
  \begin{tikzpicture}[baseline]
    \foreach \i in {1, ..., #2} {
      \pgfmathparse{\i<=#1 ? "yellow" : "gray"}
      \edef\starcolor{\pgfmathresult}
      \draw (\i*1.75ex, 0) node[name=star\i, scorestars, fill=\starcolor]  {};
   }
  \end{tikzpicture}%
}

\title{Chinese Open Instruction Generalist:\\ A Preliminary Release}

\newcommand*\samethanks[1][\value{footnote}]{\footnotemark[#1]}

\author{ 
    \small
    Ge Zhang\textsuperscript{1\space 3}\thanks{\quad The two authors contributed equally to this work.}\space, 
    Yemin Shi\textsuperscript{1}\samethanks\space, 
    Ruibo Liu\textsuperscript{1\space 4},
    Ruibin Yuan\textsuperscript{1\space 7},
    Yizhi Li\textsuperscript{1\space 2},
    Siwei Dong\textsuperscript{1}, 
    Yu Shu\textsuperscript{1},
    Zhaoqun Li\textsuperscript{1\space 5},
    Zekun Wang\textsuperscript{1\space 6}, \\
    \small
    \textbf{Chenghua Lin\textsuperscript{2},
    Wenhao Huang\textsuperscript{1}\thanks{\quad Corresponding authors.}\;,
    Jie Fu\textsuperscript{1}\samethanks} \\ 
{\small 
\textsuperscript{1} Beijing Academy of Artificial Intelligence, China
}\vspace{-0.1mm} \\ 
{\small 
\textsuperscript{2} Department of Computer Science, The University of Sheffield, UK
}\vspace{-0.1mm} \\
{\small 
\textsuperscript{3} University of Michigan Ann Arbor, USA
}\vspace{-0.1mm} \\
{\small 
\textsuperscript{4} Dartmouth University, USA
}\vspace{-0.1mm} \\
{\small 
\textsuperscript{5} Zhejiang University, China
}\vspace{-0.1mm} \\
{\small 
\textsuperscript{6} Beihang University, China
}\vspace{-0.1mm} \\
{\small 
\textsuperscript{7} Carnegie Mellon University, USA}\vspace{-0.1mm} \\
\texttt{\small
\{ymshi, whhuang, fujie\}@baai.ac.cn\textsuperscript{1}, 
}\vspace{-0.1mm} \\
\texttt{\small
gezhang@umich.edu\textsuperscript{1}
}\vspace{-0.3cm} 
}

\begin{document}

\maketitle

\begin{abstract}
Instruction tuning is widely recognized as a key technique for building generalist language models, which has attracted the attention of researchers and the public with the release of InstructGPT~\citep{ouyang2022training} and ChatGPT\footnote{\url{https://chat.openai.com/}}.
Despite impressive progress in English-oriented large-scale language models (LLMs), it is still under-explored whether English-based foundation LLMs can perform similarly on multilingual tasks compared to English tasks with well-designed instruction tuning and how we can construct the corpora needed for the tuning.

To remedy this gap, we propose the project as an attempt to create a Chinese instruction dataset by various methods adapted to the intrinsic characteristics of 4 sub-tasks. We collect around 200k Chinese instruction tuning samples, which have been manually checked to guarantee high quality. We also summarize the existing English and Chinese instruction corpora and briefly describe some potential applications of the newly constructed Chinese instruction corpora.
The resulting \textbf{C}hinese \textbf{O}pen \textbf{I}nstruction \textbf{G}eneralist (\textbf{COIG}) corpora are available in Huggingface\footnote{\url{https://huggingface.co/datasets/BAAI/COIG}} and Github\footnote{\url{https://github.com/BAAI-Zlab/COIG}}, and will be continuously updated.
% with various existing methods for performing instruction tuning on LLMs to provide helpful non-toxic Chinese instruction-tuned models to everyone.
\end{abstract}

\section{Introduction}
% research goal: 1. verify the efficiency on Chinese instruction tuning.
% 2. RLHF may not be necessary
% motivations

Pre-trained large-scale language models (LLMs) have shown revolutionary performance in many downstream tasks~\citep{hc3,flan}. One crucial ability of LLMs is called instruction following. That is, models can complete the tasks described by instructions given as input. 
%in a satisfactory way. 
This ability is based on a specialized training stage called \textit{instruction tuning}. 
Compared to unlabeled data used for pre-training, the data for instruction tuning is typically more goal-oriented, and it should explicitly demonstrate how a \textit{response} follows its corresponding \textit{instruction} with a given \textit{input}.

\begingroup
\setlength{\tabcolsep}{3pt}
\begin{table}[!t]
\centering
\resizebox{0.45\textwidth}{!}{%
\begin{tabular}{@{}lcccc@{}}
\toprule
                & Verification & Format  & Culture & Scaling   \\ \midrule
General Purpose &      \score{1}{3}      &    \score{2}{3}     &   \score{2}{3}      & \score{3}{3} \\
Academic Exams  & \score{2}{3}    & \score{2}{3} &   \score{2}{3}      &    \score{2}{3}     \\
Value Alignment &  \score{1}{3}   &  \score{1}{3}     & \score{3}{3} &      \score{1}{3}   \\
Counterfactual &  \score{2}{3}   & \score{1}{3}      &     \score{1}{3}    &    \score{2}{3}     \\
Code            & \score{3}{3}    & \score{3}{3}&     \score{1}{3}    & \score{2}{3} \\ \bottomrule
\end{tabular}%
}
\caption{The four dimensions we consider when constructing COIG instruction following data. \textbf{Verification}: whether the response can be verified. \textbf{Format}: whether the format is crucial. \textbf{Culture}: whether the response will depend on a certain culture. \textbf{Scale}: whether scaling is important. The number of filled stars presents the importance of a certain factor.}
\label{tab:overview}
\end{table}
\endgroup{}

% summarize the corpora that we have built and the major ways of building Chinese instruction corpora
There are many instruction tuning datasets in English. 
For example, the FLAN collection~\citep{flancollection} contains 15M examples %belonging to 
covering 1836 tasks, and OPT-IML~\citep{iyer2022opt} claims to have 18M examples for more than 2000 tasks (although it is still not publicly available). 
In contrast, existing data resources for Chinese instruction tuning are either small in scale or have questionable quality. 
For example, \citet{luotuo} directly translate English instruction tuning data into Chinese, but do not consider mitigating translation errors or potential cultural gaps, e.g. Chinese prefer self-sacrifice spirit while most Western countries prefer self-expression and individualistic heroism, between languages. 
%but ignore translation errors or potential cultural gaps between languages. 
For datasets that are mostly constructed through generations using Chinese LLMs \citep{belle, xu2023baize, leng2023chinese-vicuna}, they typically lacks a thorough data verification process for quality assurance. 
%but humans do not thoroughly verify the data. 
Therefore, we are motivated to develop new instruction-tuning corpora, \textbf{C}hinese \textbf{O}pen \textbf{I}nstruction \textbf{G}eneralist (\textbf{COIG}), that is larger, more diverse, and manually verified by humans.
This ensures its quality, which echoes the findings of Alpaca~\citep{alpaca} that high-quality and diverse demonstrations are crucial for good instruction-following performance.

% The translated Alpaca~\citep{alpaca, alpaca_chinese_dataset} instruction corpus is only for noncommercial use and has not been verified by humans.
% The Belle \citep{belle} corpus is large but has also not been manually verified.
 
% We release this corpus as a much larger and more diverse seed instruction set for the self-instruct~\citep{selfinstruct} procedure to generate a much larger instruction tuning corpus.
% This corpus can also be directly used for instruction tuning, as indicated by \citep{luotuo}, where its instruction set has a similar size as ours.

We highlight some unique features of COIG:

\begin{itemize}
    \item \textbf{Domain Adaption}: As shown in Tab.~\ref{tab:overview}, we consider four dimensions of instruction-tuning datasets (Verification, Format, Culture, Scaling). For each domain, we adapt our data collection pipeline to better reflect the domain specialty.
    
    \item \textbf{Diversity}: We consider a variety of tasks, including common sense reasoning, human value alignment, code generation, and hallucination correction, while very few Chinese instruction tuning data is deliberately designed for such a complete spectrum.

    \item \textbf{Quality Check by Humans}: Compared to existing model-generated Chinese instruction corpora, including \citep{luotuo,belle,instructionwild,guanaco_dataset}, COIG translated corpus is carefully verified by human annotators. 
    Moreover, since COIG translated corpus is translated from English instruction corpora~\citep{supernaturalinstructions,honovich2022unnaturalinstructions,selfinstruct} with diverse tasks, it is much more diverse than Chinese instruction corpora built by adapting prompt engineering on existing Chinese datasets, e.g. \citep{zeng2023glm-130b,Firefly,hc3}. 
\end{itemize}

% To our best knowledge, very few Chinese instruction corpora are designed for applications like common sense reasoning, human value alignment, code generation, and hallucination correction.

The main portion of COIG data is actual data that already exists on the Web, and we convert it into the proper instruction-following manner in terms of their characteristics. For example, for the academic exams domain, we crawled and manually annotated 63.5k instructions from the Chinese National College Entrance Examination, Civil Servant Examination, etc. COIG also features in including data on human value alignment in the Chinese-speaking world, and leetcode-based instruction following samples for programming. To ensure the final data quality, we hired 223 Chinese college students as quality checkers, to help us with data filtering, correction, and ratings. The resulting COIG corpus is a comprehensive set that can equip Chinese LLMs with strong instruction-following abilities in many domains.

% Second, we further sample part of the ethical exam instructions, merge them with part of verified translated \citep{hh-rlhf} as the seed instruction set, and apply the self-instruct~\citep{selfinstruct} procedure to generate a 3k corpus for Chinese human value alignment.
% We also release a subset of the exam instructions, automatically detected as related to ethics and might benefit the Chinese human value alignment.
% Finally, we propose a leetcode-based instruction corpus to help the code generation instruction-following ability of Chinese LLMs and an automatically generated multi-round Chinese Hallucination Correction Chat corpus based on Baidu Encyclopedia and ChatGLM~\citep{zeng2023glm-130b}.
% We also share some ongoing efforts and plans for corpus building in \S~\ref{sec: discussion} to encourage the whole community to collaborate with us on a better COIG corpus. 

In addition, we provide insights into the data construction pipeline based on empirical observations. We demonstrate that selecting the proper pipeline for different domains is crucial, and we have suggested the best practice for constructing instruction-tuning data in the domains COIG covers ($\S$~\ref{sec: coig}), which can be used as a reference for future instruction corpus construction workflow design.

% The paper's contribution
The paper's contributions are as follows:
\begin{itemize}
    \item To the best of our knowledge, this is one of the very first research works specifically summarizing the existing Chinese instruction tuning corpora and providing insights about how future Chinese instruction tuning corpora can be constructed.
    \item We construct 5 open-source high-quality Chinese instruction corpora, including a 68k general Chinese instruction corpus, a 62k Chinese exam instruction corpus, a 3k Chinese human-value alignment corpus, and a 13k Chinese Counterfactual Correction Multi-round Chat corpus, as samples of constructing new Chinese instruction corpora along the research directions pointed out.
    \item We construct a manually verified general high-quality Chinese instruction tuning corpus which can be directly used for Chinese LLMs' instruction tuning, both commercial and non-commercial.
\end{itemize}

\section{Existing Instruction Corpora}
\begin{table*}[htbp]
\centering \small
\caption{%English Instruction Data. ``E'' in the ``Lan'' column stands for ``English'', and ``M'' stands for ``Multilingual''. 
%``IT'' in the ``Usage'' column means ``Instruction Tuning'', and ``RL'' means ``RLHF''. 
% There are overall four data collection methods: (1)human annotated; (2)model generated; (3)existing dataset; (4)mixture of multiple methods. 
Note that the number of tasks, task types, instructions, and samples are not equivalent to one another. Only the number of tasks and instructions are reported. If the instructional data was obtained from existing public datasets and the data processing pipeline is publicly available, it is considered open-sourced. 
The field ``Verified?'' refers to whether the data has been manually verified.
%The term ``Pure QA?'' refers to whether the source data is exclusively question-answering format or not.
}
\label{tab:english-instruction-1}
\resizebox{\textwidth}{!}{
\begin{tabular}{|p{4.5cm}|p{1cm}|c|p{1.4cm}|p{1.4cm}|p{.9cm}|p{0.8cm}|p{1.2cm}|}
\toprule
\textbf{Dataset} & \textbf{\# Tasks} & \textbf{\# Instructions} & \textbf{Lan} & \textbf{Collection Method} & \textbf{Usage} & \textbf{Access} & \textbf{Verified?} \\
\hline
 PromptSource \citep{bach2022promptsource} & 180 &  2,085 & English & Mixed & Instruct. Tuning & Open & Yes \\
 \hline
  P3~\citep{p3} & 270 & 2,073 & English & Mixed & Instruct. Tuning & Open & Yes \\
 \hline
  xP3 \citep{xp3} & 83 & - & Multilingual & Mixed & Instruct. Tuning & Open & No \\
   \hline
  Natural Instruct v1 ~\citep{naturalinstructions} & 61 & 61 & English & Existing & Instruct. Tuning & Open & Yes \\
   \hline
  Super-Natural-Instruct v2 \citep{supernaturalinstructions} & 1,616 & 1,616 & Multilingual & Mixed & Instruct. Tuning & Open & Yes \\
   \hline
  CrossFit~\citep{ye-etal-2021-crossfit} & 160 & - & English & Existing & Instruct. Tuning & Open & Yes \\
   \hline
  FLAN~\citep{flan} 2021 & 62 & 620 & English & Existing & Instruct. Tuning & Open & Yes \\
   \hline
  ExMix~\citep{aribandi2021ext5} & 107 & 107 & English & Existing & Instruct. Tuning & Open & - \\
   \hline
  UnifiedSKG~\citep{unifiedskg} & 21 & 21 & English & Existing & Instruct. Tuning & Open & Yes \\
   \hline
  MetaICL~\citep{min2021metaicl} & 142 & - & English & Existing & Instruct. Tuning & Open & Yes \\
   \hline
  InstructGPT \citep{ouyang2022training} & - & 112,801 & English & Human Annotated & RLHF, Instruct. Tuning & Closed & Yes \\
   \hline
  FLAN Collection 2022 \citep{flan2022, flancollection} & 1,836 & 18,360 & English & Existing & Instruct. Tuning & Open & No \\
   \hline
  OPT-IML Bench      ~\citep{iyer2022optiml} & 1,667 & 3,128 & English & Existing & Instruct. Tuning & Open & Yes \\
   \hline
  GLM-130B~\citep{zeng2023glm-130b} & 74 & - & Multilingual & Existing & Instruct. Tuning & Open & Yes \\
   \hline
  Self-Instruct~\citep{selfinstruct} & 175 & 52,445 & English & Model Generated & Instruct. Tuning & Open & No \\
   \hline
  Unnatural Instructions~\citep{honovich2022unnaturalinstructions} & - & 240,000 & English & Model Generated & Instruct. Tuning & Open & No \\
   \hline
  Alpaca~\citep{alpaca} & 175 & 51,942 & English & Model Generated & Instruct. Tuning & Open & No \\
\bottomrule
\end{tabular}
}
\end{table*}

\begin{table*}[htbp]
\centering \small
\caption{English Instruction Data (Continued from Table \ref{tab:english-instruction-1})}
\label{tab:english-instruction-2} 
\resizebox{\textwidth}{!}{
\begin{tabular}{|p{4.5cm}|p{1cm}|c|p{1.4cm}|p{1.4cm}|p{1.5cm}|p{0.8cm}|p{1.2cm}|}
\toprule
\textbf{Dataset} & \textbf{\# Tasks} & \textbf{\# Instructions} & \textbf{Lan} & \textbf{Collection Method} & \textbf{Usage} & \textbf{Access} & \textbf{Human Verified?} \\
 \hline
  OIG \citep{oig_dataset} & 30 & ~43M & English & Mixed & Instruct. Tuning & Open & No \\
   \hline
  Baize~\citep{xu2023baize} & 3 & 100K+ & English & Model Generated & Chat & Open & No \\
\hline
  Camel~\citep{camel} & - & 115K & English & Model Generated & Instruct. Tuning, Chat & Open & No \\
   \hline
  UltraChat ~\citep{UltraChat} & - & 675K & English & Model Generated & Chat & Open & No \\
   \hline
  Dolly~\citep{dolly} & 7 & 15,000 & English & Human Annotated & Instruct. Tuning & Open & Yes \\
   \hline
  Guanaco-Dataset            ~\citep{guanaco_dataset} & 175 & 534,530 & Multilingual & Mixed & Instruct. Tuning & Open & No \\
   \hline
ChatLLaMA  Chinese-ChatLLaMA  ~\citep{chinese_chatllama} & - & - & Multilingual & Mixed & Instruct. Tuning & Open & No \\
   \hline
  GPT-4-LLM ~\citep{gpt4-llm} & 175 & 165K & Multilingual & Model Generated & RLHF, Instruct. Tuning & Open & No \\
   \hline
  ShareGPT ~\citep{sharegpt} & - & - & Multilingual & Model Generated & Instruct. Tuning, Chat & Closed & Yes \\
   % \midrule
  % Vicuna & - & - & English & Model Generated & Instruct. Tuning & Closed & No \\
   \hline
  SHP~\citep{stanfordhumanp} & 18 & 385K & English & Existing, Human Annotated & RLHF, Instruct. Tuning & Open & Yes \\
   \hline
  HH-RLHF \citep{hh-rlhf, hh_rlhf, red-teaming} & - & 169,550 & English & Mixed & RLHF, Instruct. Tuning & Open & Yes \\
   \hline
  HC3~\citep{hc3} & 12 & 37,175 & Multilingual & Mixed & Instruct. Tuning & Open & Yes \\
   \hline
  Stack-Exchange-Preferences~\citep{h4stackexchange} & - & 10M & English & Existing & RLHF, Instruct. Tuning & Open & Yes \\
  \hline
  InstructWild~\citep{instructionwild} & 429 & 104K & Multilingual & Model Generated & Instruct. Tuning & Open & No \\
\bottomrule
\end{tabular}
}
\end{table*}

\begin{table*}[htbp]
\centering \small
\caption{Chinese Instruction Data (Continued from Table \ref{tab:english-instruction-2})}
\label{tab:chinese-instruction-3} 
\begin{tabular}{|p{2.5cm}|p{1cm}|c|p{1.8cm}|p{1.4cm}|p{1.5cm}|p{0.8cm}|p{1.4cm}|}
\toprule
\textbf{Dataset} & \textbf{\#Tasks} & \textbf{\# Instructions} & \textbf{Language} & \textbf{Collection Method} & \textbf{Usage} & \textbf{Access} & \textbf{Human Verified?} \\
 \hline
  xP3~\citep{xp3} & 83 & - & Multilingual & Mixed & Instruct. Tuning & Open & No \\
    \hline
  Super-Natural-Instructions (v2)~\citep{supernaturalinstructions} & 1,616 & 1,616 & Multilingual & Mixed & Instruct. Tuning & Open & Yes \\
    \hline
  ZeroPrompt  ~\citep{xu-etal-2022-zeroprompt} & 1,110 & - & Chinese & Human Annotated, Existing Dataset & Instruct. Tuning & Closed & Yes \\
    \hline
  GLM-130B~\citep{zeng2023glm-130b} & 74 & - & Multilingual (eng, zh) & Existing Dataset & Instruct. Tuning & Open & Yes \\
    \hline
  pCLUE~\citep{pclue} & 9 & 73 & Chinese & Existing Dataset & Instruct. Tuning & Open & Yes \\
    \hline
  Belle-1.5M~\citep{belle} & 175 & 1.5M & Chinese & Model Generated & Instruct. Tuning & Open & No \\
    \hline
  Guanaco-Dataset    ~\citep{guanaco_dataset} & 175 & 534,530 & Multilingual & Mixed & Instruct. Tuning & Open & No \\
    \hline
  CSL~\citep{li-etal-2022-csl} & 4 & 396,209 & Chinese & Existing Dataset & Instruct. Tuning & Open & Yes \\
    \hline
  Chinese-ChatLLaMA   ~\citep{chinese_chatllama} & - & - & Multilingual & Mixed & Instruct. Tuning & Open & No \\
    \hline
  Firefly~\citep{Firefly} & 23 & 1.1M & Chinese & Mixed & Instruct. Tuning & Open & Yes \\
    \hline
  Luotuo~\citep{luotuo} & 175 & 51,672 & Chinese & Mixed & Instruct. Tuning & Open & No \\
    \hline
  Chinese-Alpaca     ~\citep{alpaca_chinese_dataset} & - & - & Chinese & Existing Dataset, Human Annotated & Instruct. Tuning & Open & No \\
    \hline
  GPT-4-LLM ~\citep{gpt4-llm} & 175 & 165K & Multilingual (eng, zh) & Model Generated & RLHF, Instruct. Tuning & Open & No \\
    \hline
  ShareGPT  ~\citep{sharegpt} & - & - & Multilingual & Model Generated & Instruct. Tuning, Chat & Closed & Yes \\
    \hline
  Chinese-Vicuna   ~\citep{leng2023chinese-vicuna} & - & 1M & Chinese & Model Generated, Existing Dataset & Instruct. Tuning & Open & No \\
    \hline
  CUGE~\citep{yao2021cuge} & 18 & - & Chinese & Existing Dataset & Instruct. Tuning & Open & Yes \\
    \hline
  HC3~\citep{hc3} & 12 & 37,175 & Multilingual (eng, zh) & Mixed & Instruct. Tuning & Open & Yes \\
    \hline
  InstructWild ~\citep{instructionwild} & 429 & 104K & Multilingual (eng, zh) & Model Generated & Instruct. Tuning & Open & No \\ \hline
  Our Translated Corpus & 2k & 67,798 & Chinese & Mixed & Instruct. Tuning & Open & Yes \\
\bottomrule
\end{tabular}
\end{table*}

Large language models (LLMs) fine-tuned to respond to specific instructions have demonstrated a remarkable zero-shot ability to generalize to new tasks. One key ingredient is the curation of the instruction data, for which the research community has developed various strategies for dataset construction. 
In this section, we provide a comprehensive summary of the English and Chinese instruction corpora in Tab.~\ref{tab:english-instruction-1}, \ref{tab:english-instruction-2} and Tab.~\ref{tab:chinese-instruction-3}, respectively. We also describe the mainstream approaches for constructing the instruction-tuning datasets below. 

\noindent\textbf{Human annotation.}~~Early attempts to construct instruction data is typically through human annotation~\citep{naturalinstructions,supernaturalinstructions,dolly}. Representative works include PromptSource~\citep{bach2022promptsource} and Super-Natural-Instructions~\citep{supernaturalinstructions}, both instructional datasets that require extensive manual/expert annotations to collect instructions that can train models to follow various in-context instructions. Although human-annotated instruction data is generally of high quality, they are also limited in quantity, diversity, and creativity. This limitation has a significant impact. Previous research has demonstrated a direct correlation between the size and diversity of instructional data and the generalizability of the resultant models to previously unseen tasks~\citep{selfinstruct}.

\noindent\textbf{Semi- and automatic construction.}~~To address this bottleneck and reduce dependence on human annotators, researchers have proposed various methods, ranging from semi-automatic~\citep{selfinstruct}, to fully automatic instruction generation~\citep{honovich2022unnaturalinstructions}. 
Self-Instruct~\citep{selfinstruct} is a bootstrapping framework that utilizes an initial set of manually-written instructions to guide the expansion of the instructions. 
The framework generates its instructions and aligns its outputs with them, resulting in enhanced instruction-following abilities of LLMs. Motivated by recent research on leveraging language models for data generation, \citet{honovich2022unnaturalinstructions} propose to collect instructions by prompting an LLM. This involves eliciting additional instruction examples using a limited number of seed instructions, and further expanding the dataset by soliciting the model to rephrase each instruction. To promote creativity, stochastic decoding is utilized to generate diverse example inputs, while deterministic decoding is employed for output generation to ensure accuracy. 
%for numerous downstream tasks. 
 % [CL] Cons: noise introduced during the process; 
Although automated or semi-automated methods for data construction significantly reduce the need for human labor, they may also result in a substantial amount of noise in the generated samples. For instance, Unnatural Instructions can exhibit up to 50\% noisy samples. Therefore, it is crucial to implement mechanisms (e.g., data pruning) that can mitigate this challenge and improve the usability of the data. 
%\textbf{[TODO: to describe a few more relevant approaches]]}

\noindent\textbf{LLM Society.}~~
Given a set of manual configuration, communicative or generative agents represent a promising alternative approach to generate instruction or chat corpora~\citep{camel,xu2023baize,park2023generative}.
\citet{camel} claim that the instruction and chat corpora generated by communicative agents can retain many useful characteristics of CoTs and self-refinement.
\citet{xu2023baize} validate that
 self-chat corpus generated by only giving ChatGPT manual configurations is helpful for aligning LLM with human preferences.
%\noindent\textbf{Prompting.}~~

\noindent\textbf{Translation.}~~In addition to the aforementioned methods, translation is also a primary method for constructing Chinese instruction tuning corpora, that is, translating English corpora into Chinese. 
%Existing translated Chinese instruction tuning corpora are majorly constructed by MT engines with little manual verification, e.g. Luotuo~\citep{luotuo}.
Representative datasets constructed through translation include Luotuo~\citep{luotuo}, BELLE~\citep{belle}, and Alpaca~\citep{alpaca_chinese_dataset}, which are constructed by machine translation engines with little or no manual verification.
%or via model generation, e.g., \citep{belle, xu2023baize, leng2023chinese-vicuna}. 
Furthermore, the popular Alpaca~\citep{alpaca, alpaca_chinese_dataset} instruction corpus is limited to noncommercial use only. 
Nevertheless, the Chinese instruction corpora are much scarcer, compared to the English corpora available. 
%and has not been verified by humans.
%The BELLE~\citep{belle} corpus is in a considerable size but has not been manually verified as well.

\section{COIG: Chinese Open Instruction Generalist}
\label{sec: coig}
To address the scarcity of instruction corpora, we propose the Chinese Open Instruction Generalist (\textbf{COIG}) project to maintain a harmless, helpful, and diverse set of Chinese instruction corpora.
We welcome all researchers in the community to contribute to the corpus set and collaborate with us.
We only release the first chip of COIG to help the Chinese LLMs' development in the exploration stage and appeal to more researchers joining us in building COIG. 
We separately introduce a manually verified translated general instruction corpus in \S~\ref{sec: translated}, a manually annotated exam instruction corpus in \S~\ref{sec: exam}, a human value alignment instruction corpus in \S~\ref{sec: human value alignment}, a multi-round counterfactual correction chat corpus in \S~\ref{sec: human value alignment}, and a leetcode instruction corpus in \S~\ref{sec: leetcode}.
We provide these new instruction corpora to assist the community with instruction tuning on Chinese LLMs.
These instruction corpora are also template workflows for how new Chinese instruction corpora can be built and expanded effectively.

\subsection{Translation-based General Instruction Corpus}
%\subsection{Translated Instructions}
\label{sec: translated}
To enable the corpus for commercial and non-commercial use, we carefully select the core data of unnatural instructions~\citep{honovich2022unnaturalinstructions}, the seed instruction set of self-instruct~\citep{selfinstruct}, and task descriptions of supernatural instructions~\citep{supernaturalinstructions} as the English instruction source.
These source instructions are not generated by any OpenAI API, and are therefore available for commercial and non-commercial use~\citep{selfinstruct,honovich2022unnaturalinstructions,supernaturalinstructions}.
There are 67798 instructions in total, which are composed of 1616 task descriptions in \citep{supernaturalinstructions} along with a single instance for each of them, 175 seed tasks in \citep{selfinstruct}, and 66007 instructions from \citep{honovich2022unnaturalinstructions}.

To reduce the cost and further improve the quality of the instruction corpus, we separate the translation procedure into three phases: automatic translation, manual verification, and manual correction.
\textbf{First}, during the automatic translation phase, we concatenate the instruction with the input and output of the instances and feed them into DeepL\footnote{\url{https://www.deepl.com/translator}} for translation.

\textbf{Second}, during the manual verification phase, we define 4 labels for the annotators to select for each instruction.
Each instruction is (i) directly usable; (ii) usable but with the source input and output of the instance; (iii) usable but with manual correction, and (iv) not usable.
There are fewer than twenty cases where it is not usable.
We adopt a two-phase quality verification for the manual verification phase:
In the first phase, each case is verified by an industrial experienced quality inspector with more than 5 years of work experience after being annotated by the annotator.
The entire corpus can be passed into the second quality verification phase if and only if the correctness rate exceeds $95\%$.
The corpus got a $96.63\%$ correctness rate in the first quality verification phase in the final.
Our expert quality inspectors (namely, our coauthors) are in charge of the second quality verification phase and only randomly sample 200 cases from the total corpus for quality verification.
If and only if all sampled cases are classified correctly, the corpus is able to be passed into the manual correction phase.

\textbf{Third}, during the manual correction phase, the annotators are asked to correct translated instructions and instances into correct Chinese \{instruction, input, output\} triplets instead of just keeping the translation correct. 
The annotators are asked to do it because there exist factual errors in source unnatural instructions which might lead to LLMs' hallucinations.
There are 18074 instructions fed into the manual correction phase in total. 
We use the same two-phase quality verification procedure as the manual verification phase.
The corpus got a $97.24\%$ correctness rate in the first quality verification phase of the manual correction phase.

These strict quality verification procedures assure the reliability of the translated corpus.

\subsection{Exam Instructions}
\label{sec: exam}
\begin{figure}[t]
    \centering
    \includegraphics[width=7.5cm]{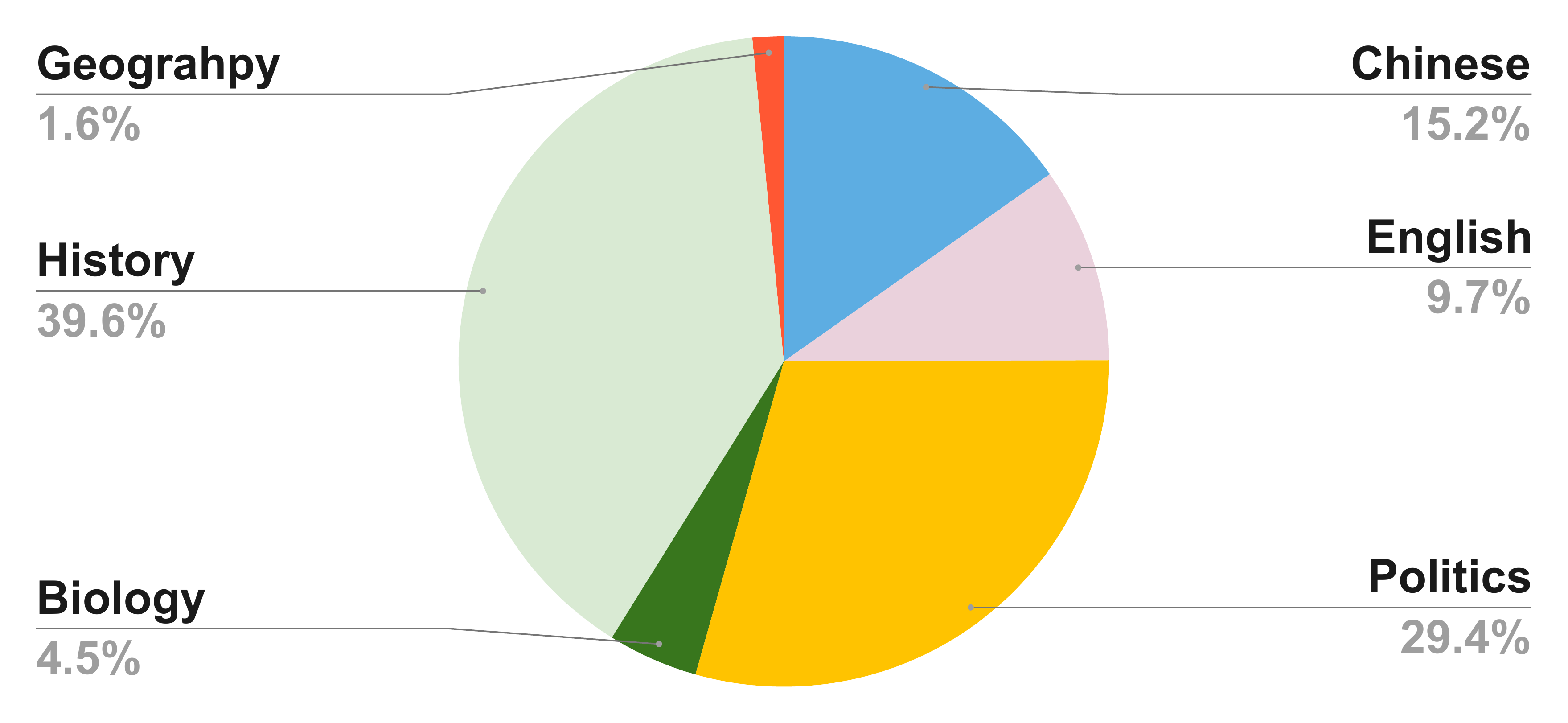}
    \caption{The percentage of instructions of different coarse-grained subjects.}
    \label{fig: subjects}
    \vspace{-2mm}
\end{figure}

\begin{figure}[t]
    \centering
    \includegraphics[width=7.5cm]{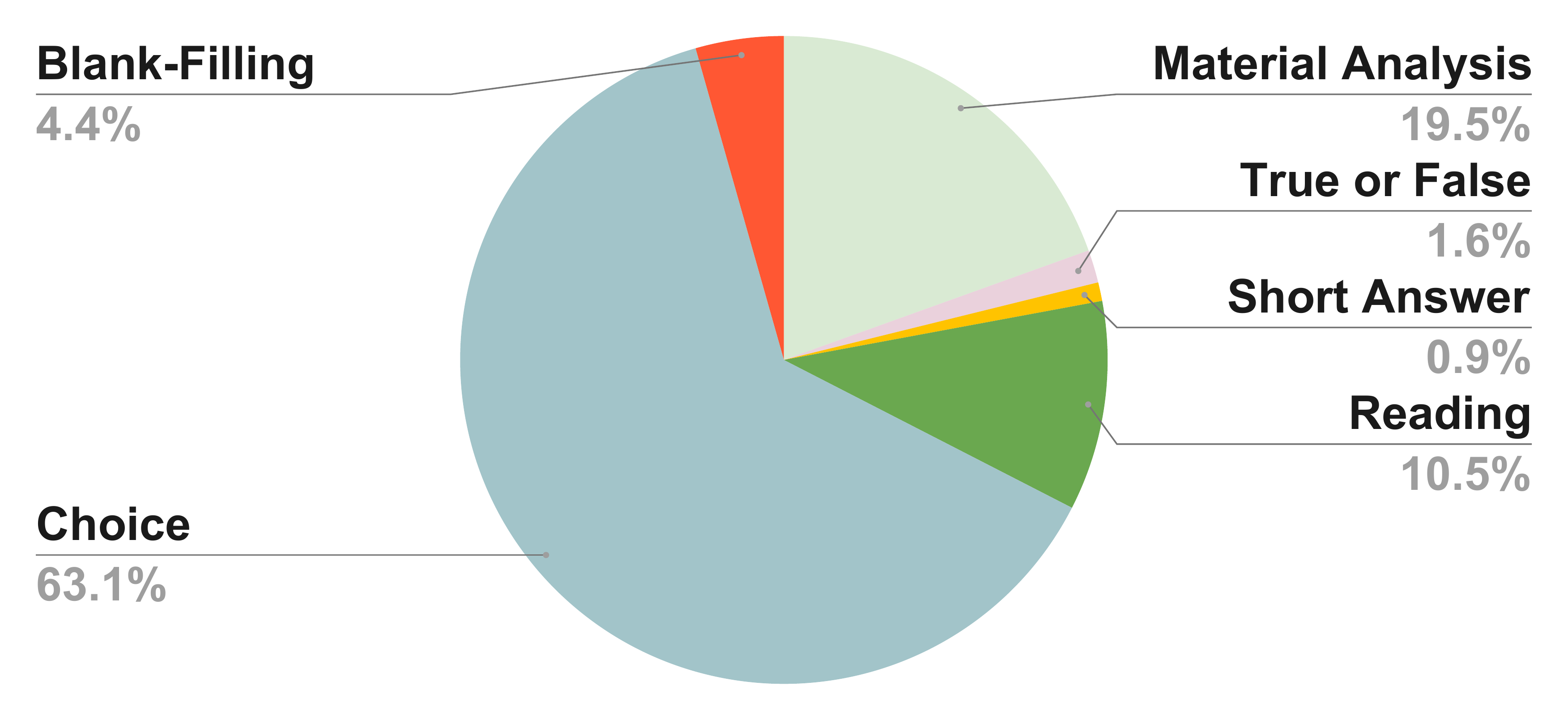}
    \caption{The percentage of instructions in different question formats.}
    \label{fig: questionFormat}
    \vspace{-2mm}
\end{figure}

The Chinese National College Entrance Examination, Middle School Entrance Examinations, and Civil Servant Examination are the main Chinese commonsense tests.
These exams contain various question formats and detailed analysis that can be used as the Chain-of-Thought (\textbf{CoT}) corpus.
We use potato~\citep{pei2022potato}, an active learning powered open-source annotation website template, for manual annotation, which extracts six informative elements from original exam questions, including instruction, question context, question, answer, answer analysis, and coarse-grained subject.
There are many reading comprehension questions in these exams, and the question context means the reading material of these reading comprehension questions.
There are six main coarse-grained subjects: Chinese, English, Politics, Biology, History, and Geology.
There are very few Math, Physics, and Chemistry questions in the corpus because these questions are often with complex symbols which are hard to annotate.
We illustrate the question format percentage in Fig.~\ref{fig: subjects} and the major subject percentage in Fig.~\ref{fig: questionFormat}.
For many choice questions, we recommend that the researchers utilize this corpus to further post-process it using prompts or post-process it to blank-filling questions to increase the instructions' diversity further.

\subsection{Human Value Alignment Instructions}
\label{sec: human value alignment}
Many existing human value alignment datasets can serve instruction tuning~\citep{hh_rlhf,forbes2020social,emelin-etal-2021-moral}; however, these datasets are in English, and we find that simply translating them into Chinese cannot produce high-quality alignment data that matches the unique culture rooted in the Chinese-speaking world. For example, in Western English-speaking countries, people are often encouraged to move out when they hit adulthood, while in many Chinese-speaking communities or broader East Asian countries, it is acceptable or even encouraged for the youth to still live with and accompany their family even though they have grown up. 

To respect and reflect this major difference caused by different cultural backgrounds, different from other tasks in COIG that leverage one unified collection of instruction-following samples, we categorize the value alignment data into two separate sets: 1) a set of samples that present shared human values in the Chinese-speaking world, and 2) some additional sets of samples that present regional-culture or country-specific human values. For the first shared set, we choose self-instruct~\citep{selfinstruct} as the main method to augment a set of seed instruction-following samples. For the additional sets, to guarantee that the data genuinely 
%honestly
reflect the local values, we mainly rely on web crawlers to collect the data in original forms.

The seed instructions for shared human values are manually picked from Chinese textbooks and exams on ethics education, since we believe most of the content in these materials has already considered the common ground of different communities (e.g., there are 56 minorities in China). We deliberately consider the following three principles when filtering the data:

\begin{itemize}
    \item It should present shared human values widely accepted in the Chinese-speaking world, rather than regional ones.
    \item It should not include political propaganda or religious beliefs and should not be related to disputed claims.
    \item It should not just explain proverbs or quotes, since they will likely be covered in the knowledge retrieval instructions-following data.  
\end{itemize}

In total, we choose 50 instructions as the augmentation seeds, and produce 3k resulting instructions following samples for general-purpose value alignment in the Chinese-speaking world. Meanwhile, we also collect 19,470 samples as a regional addition, which is specific to users in China (including many terms that are only used in the Chinese community). See \ref{sec:HVAI} for examples.

\subsection{Counterfactual Correction Multi-round Chat}
\label{sec: counterfactual}
LLMs have become ubiquitous in a variety of NLP applications. However, these models often generate responses that are not truthful, and in some cases can even propagate misinformation or hallucination. The models may falsely and repeatedly insist a claim with no sign of internal awareness that the claim was a product of their own imagination. 

To mitigate above issues, and enhance the truthfulness of the model's responses, we build the Counterfactual Correction Multi-round Chat dataset (CCMC). It is constructed based on the CN-DBpedia knowledge base \citep{xu2017cn} with the aim of alleviating and resolving the pain points of hallucination and factual inconsistency 
%disrespect 
in current LLMs. The original knowledge base consists of 5634k entities with their corresponding attribute-value pairs and original text.

The CCMC dataset includes 5 rounds of role-playing chat between a student and a teacher, and the corresponding knowledge they refer to. The teacher generates responses based on ground-truth knowledge and corrects factual errors or inconsistencies in the student's questions or statements in each round. 
In the final round, the teacher will summarize the chat and review the confusing terms, i.e. the factual errors or inconsistencies in the student's questions or statements. 
The dataset contains 13,653 dialogues and resulting in 68,265 rounds of chat.
See \ref{sec:CCMC} for example.

We outline the workflow for creating the CCMC dataset. The workflow consists of three main parts: entity selection, information extraction, and chat generation.

We first perform entity selection by ranking entities based on entity tag frequency and choosing the top 200. 
We prioritize entities with summaries and aim to retain factual/knowledge-based content, such as well-established, historically tested concepts, and entities related to various academic disciplines, historical events, and social events. 
Tags like organizations, companies, foods, and games are excluded.

Next, we extract information from the knowledge base using a chat LLM. 
We first obtain a source entity by randomly sampling an entity from the high-priority categories, returning triplets, content summaries, and content section titles. Then we ask a chat LLM to summarize all the information into a better summary and also extract attribute-value pairs from the input. 
This can filter out some of the false tags in the Baidu tags, and also take information in the unstructured content into consideration. 
For the confusion entity, we use a prompt-based method to extract a list of confusing terms based on the input information. Then we match the terms with the knowledge base. 
If the term exists in the base, we keep the term and use the same method to extract better summary and attribute-value pairs.

We employ a teacher-student question-and-answer approach for chat generation to generate attack and defense scenarios gradually. 
We provide the extracted original entity summary and confusing entity summary. Then, we let the student ask the teacher about the original concept while mistakenly mixing it up with the confusing one. The teacher would then clarify and differentiate the concepts in a JSON format. 
The conversation would continue for multiple rounds, each time with the student challenging the teacher based on previous dialogues, and the teacher providing clarifications and distinctions. 
In the final round, the teacher would reintroduce the original concept and summarize the concepts that were easily confused, emphasizing and differentiating the concepts the student had previously mixed up. 
All chats are generated by prompting a chat LLM.

\subsection{Leetcode Instructions}
\label{sec: leetcode}
\begingroup
\setlength{\tabcolsep}{2pt}

\begin{table}[ht]
\centering
\caption{Statistics of Leetcode Instructions. Task types C2T and T2C refer to code-to-text and text-to-code, respectively. And \textit{e.} is the abbreviation for ``explanation''. Programming languages with less than 50 instructions are merged into the ``Others'' class.}
\label{table:leetcode_stats}
\resizebox{0.5\textwidth}{!}{%
\begin{tabular}{lccccc}
\toprule

\multirow{2}{*}{
\textbf{\makecell[c]{Programming  \\ Language}
}} & \multicolumn{4}{c}{\textbf{Task Type}}\\
 & \makecell[c]{C2T w/o e.} & \makecell[c]{C2T w/ e.} & \makecell[c]{T2C w/o e.} & \makecell[c]{T2C w/ e.} & \makecell[c]{All} \\
\midrule
C & 8 & 76 & 12 & 89 & 185 \\
C\# & 8 & 56 & 6 & 58 & 128 \\
C++ & 168 & 943 & 180 & 963 & 2254 \\ 
Go & 175 & 1008 & 164 & 899 & 2246 \\ 
Java & 213 & 989 & 193 & 983 & 2378 \\ 
JavaScript & 16 & 172 & 29 & 153 & 370 \\ 
Python3 & 198 & 995 & 208 & 981 & 2382 \\ 
Rust & 46 & 252 & 39 & 252 & 589 \\ 
SQL & 35 & 6 & 30 & 5 & 76 \\ 
TypeScript & 98 & 454 & 82 & 450 & 1084 \\ 
Others & 2 & 20 & 3 & 20 & 45 \\ 
\midrule
All & 967 & 4971 & 946 & 4853 & 11737 \\
\bottomrule
\end{tabular}
}
\end{table}

\endgroup

% intuition
Given that the code-related tasks potentially contribute to the ability emergence of LLMs~\citep{ouyang2022training}, we argue that code-related tasks aligned with the Chinese natural language should be considered in our datasets.
Therefore, we build the Leetcode instructions from a \textit{CC-BY-SA-4.0} license collection\footnote{\url{https://github.com/doocs/leetcode}} of $2,589$ programming questions. The questions contain problem descriptions, multiple programming languages, and explanations\footnote{$834$ questions do not have explanations.}.

% our coversion method
We categorize the instruction tasks into two classes considering the input and output: code-to-text and text-to-code. 
The code-to-text task requires producing function descriptions given programming codes, whereas the text-to-code task requires output codes from the question. 
Depending on whether the program question has a corresponding explanation, the task instruction will be distinguished by \textit{with/without explanation}.
We prepare $38$ types of descriptions to generate the Leetcode instructions. We iterate through the available programming language implementations for each programming question, randomly sample the task as code-to-text or text-to-code, and then randomly select a corresponding instruction description.

% summary
The statistics of the Leetcode instructions are shown in Tab.~\ref{table:leetcode_stats}. 
We derive $11,737$ code-related instructions in four types of tasks in more than $10$ programming languages from the collection. The statistics show that the constructed dataset is diversified and may benefit LLM instruction tuning.

\subsection{Empirical Validation of Instruction Corpora Construction Workflow}
This section summarizes reasonable empirical conclusions and lessons about the Chinese instruction corpora construction workflow.

\textbf{First}, adopting In-Context-Learning (\textbf{ICL}) for generating new instructions~\citep{selfinstruct,honovich2022unnaturalinstructions} is a key contributing factor when we want to expand the size of the instruction corpus.
Taking the general-purpose instruction corpora~\citep{belle, alpaca} in Tab.~\ref{tab:overview} as examples, generating these instructions is more realistic using the ICL ability of existing LLMs instead of relying on manual annotation or other methods\footnote{GLM-130B~\citep{zeng2023glm-130b}, T5~\citep{raffel2020exploring}, and various other LLMs are also capable of performing the ICL procedure needed to generate new instructions~\citep{honovich2022unnaturalinstructions, selfinstruct} but with a relatively higher rate of error compared to ChatGPT.}. 
LLM developers should carefully decide which LLMs and seed instruction corpora they prefer based on the license of the source, the relationship of the source with OpenAI\footnote{OpenAI does not allow the content generated from their applications to be used to improve model performance without permission. For details, see \url{https://openai.com/policies/terms-of-use}.}, and their needs.

\textbf{Second},  human annotation or verification is needed when there is a cultural difference between the targeted language and the language of source instruction corpora.
As in \S~\ref{sec: human value alignment}, we must carefully select the seed in manual instruction to ensure that the seed instructions align well with Chinese culture and do not include political propaganda or regional beliefs.
We also recommend using existing corpora, such as the method introduced in \citep{stanfordhumanp} when building human value alignment instructions, where one crawls corpus from forums and post-processes it to make it harmless.

\textbf{Third}, model-generated corpora need more detailed manual quality verification, especially in cases where the output format is crucial.
During the translation and verification procedure of the unnatural instructions~\citep{honovich2022unnaturalinstructions} explained in \S~\ref{sec: translated}, we notice many instances that do not follow the model-generated instructions and a considerable number of imperfect model-generated instructions.
Another concern is that the diversity and distribution of model-generated instructions are highly dependent on seed instructions.
Manual selection and verification may help sample an instruction corpus from a large raw instruction corpus with a more balanced distribution and better diversity than the large raw instruction corpus itself, as indicated in \citep{koala_blogpost_2023}.
\section{Conclusion and Discussion}
\label{sec: discussion}

We have described how we build the most comprehensive Chinese instruction dataset with careful human verification. 
Since our aim is to build a community based on the continuous update philosophy, this early-phase release is a solid foundation and momentum for future evolution and improvement. 
Note that ``early-phase'' does not imply that the current version is highly incomplete, but emphasizes that we commit to updating the corpora and welcome contributions from the community. 
Our next major release will focus on building better instructions for collective LLMs~\citep{park2023generative,camel,xu2023baize} and how to use them to improve the training of LLMs in return. 

In addition to our efforts in constructing the instruction corpora, we also want to share our thoughts on several potential algorithmic improvements that warrant investigation. 
For example, it is well-known that not every training sample is equally useful when training machine learning models. 
Thus, it is reasonable to conjecture that some instructions may be more important than others when fine-tuning LLMs. 
In practice, we are always constrained by the limited resources for a specific training task. 
Fine-tuning LLMs with redundant instructions may give little benefit, and constructing meaningful and high-quality instructions is difficult and costly. 

Furthermore, inspired by the detrimental gradient interference phenomena studied in~\citep{yu2020gradient}, we hypothesize that the diverse nature of instructions could make fine-tuning challenging as the gradients\footnote{It becomes even more tricky when we pack these instructions into a mini-batch, usually constrained by the on-device memory capacity. One remedy is to design the corresponding gradient accumulation scheme with care. } obtained by various instructions might even conflict with each other, making the optimization process difficult to converge to good solutions.
One solution is to adopt active learning approaches to proactively and iteratively construct the most informative instructions that benefit LLM the most, rather than creating them simultaneously and treating them equally important. 
This active learning-based approach requires a set of comprehensive evaluation and diagnosis toolkits to guide the active construction procedure of instructions.  
Additionally, humans learn much better and faster if the training samples are presented in a meaningful order which usually exposes the learners to gradually increasing complexity \citep{bengio2009curriculum}. 
It makes sense to incorporate this curriculum learning principle~\citep{wang2021survey} into the active learning framework~\citep{jafarpour2021active}. 
Alternatively, after generating these instructions, we can learn to reweight the instructions within a mini-batch or across multiple mini-batches, using meta learning~\citep{ren2018learning} to mitigate the issue of conflicting gradients. 

% Another related issue is catastrophic forgetting~\citep{goodfellow2013empirical,kirkpatrick2017overcoming}, where LLMs are prone to ``forgetting'' what they have learned from one specific set of instructions after being exposed to another set of instructions. 

\section{Acknowledgement}

We would like to express our gratitude to the student volunteers from Zhejiang University, China University of Geosciences (Beijing), Beijing Language and Culture University, Zhengzhou University, Beijing University of Posts and Telecommunications, Beihang University, and Hangzhou Dianzi University for their manual annotation and quality control efforts. We also appreciate the non-profit annotation services provided by Stardust.ai in the construction of the translation dataset.

\bibliography{custom}
\bibliographystyle{apalike}

\newpage

\appendix

\section{Appendix}
\label{sec:appendix}

\begin{CJK*}{UTF8}{gbsn}
\subsection{Human Value Alignment Instructions}\label{sec:HVAI}
\begin{table}[htbp]
\centering
\scalebox{0.85}{
\begin{tabular}{|p{0.95\linewidth}|}
\hline
\small{\textbf{Instructions Sample}}\\ \hline
\small{
Instruction:

假如你是一名校长，最近接到不少学生和家长反映，学校操场上某些同学总是进行踢足球、打篮球等活动，严重影响了其他同学的学习和休息。请你制定一份公告，呼吁全体学生遵守校规，有序使用学校场地。

}\\ \midrule
\small{
Input: 

公告起草委员会已经开会商讨该公告的内容，加入你的建议，你需要在公告中提供具体制度，呼吁全体学生遵守校规，保证课堂和自习室的安静和干净。

}\\ \midrule
\small{
Output: 

全校同学在使用学校的场地设施，应自觉遵守学校有关规定和管理制度，保证场地的有序和安全，维护学校正常的教学、体育训练、课外活动等各项工作的进行。学校操场是同学们锻炼身体的场所，但也是学习的场所，应该保持安静，不影响其他同学的学习和休息。进入球场、相邻广场玩耍一律需老师主持，并根据实际情况合理规划使用时间；球鞋等装备必须符合安全要求。严禁未经校方批准私自使用学校未开放的区域及相关设施；公共体育设施的使用，要坚持先预约后使用的原则。我们希望同学们在学校的各个场地内，能够自律自觉，有序文明的使用，并与教职工一起共同创建良好的校园环境。
}\\ \bottomrule
\end{tabular}
}
\caption{Instructions for aligning AI with human value.}
\end{table}

\begin{table}[htbp]
\centering
\scalebox{0.85}{
\begin{tabular}{|p{0.95\linewidth}|}
\hline
\small{\textbf{Instructions Sample}}\\ \hline
\small{
Instruction:

假如您在繁忙的工作中感到压力，下列哪些做法是正确的？

}\\ \midrule
\small{
Input: 

A．乐观开朗心态，正确认识工作 

B．工作之余利用假期适当休息 

C．用唾液测试压力荷尔蒙 

D．抱怨工作压力

}\\ \midrule
\small{
Output: 

压力是生活、工作中难以避免的，如何正确应对压力是每一个人都要面对的现实问题。

AB: 首先，要保持乐观的心态，正确对待工作压力，认真思考解决问题的方法，适当调整工作、生活的节奏和方式，保持身心愉悦；当然，也需要适当的休息，放松身心，以便更好地应对工作和压力。因此，AB是正确的做法，符合题意；

C：常规唾液测试并不能测试压力，不符合题意；

D：抱怨只会让负面情绪更加放大，不利于缓解压力。所以，D不符合题意。

故本题选AB。
}\\ \bottomrule
\end{tabular}
}
\caption{Instructions for aligning AI with human value.}
\end{table}

\newpage
\subsection{Counterfactual Correction Multi-round Chat}\label{sec:CCMC}
\begin{table}[htbp]
\centering
\scalebox{0.75}{
\begin{tabular}{|p{1.10\linewidth}|}
\hline
\small{\textbf{Multi-round Chat Sample}}\\ \hline
\small{
Premise:

刻点小食蚜蝇（Paragus tibialis Fallen）是双翅目昆虫纲食蚜蝇科的一种捕食性天敌，属于动物界节肢动物门。它会对棉花等作物造成危害，以蚜虫为寄主。该蚜蝇的成虫在4月份开始在绿肥、蚕豆和榆、槐、柳、木槿以及杂草等植物上产卵繁殖，并于5月上旬迁入棉田，发生数量较少。卵期约3-4天，幼虫期8-9天，蛹期12-14天，由卵到成虫历期23-27天，在6-7月期间卵期一般2-3天，幼虫期6-7天，蛹期6-7天，由卵到成虫历期14-17天。
}\\ \midrule
\small{
Confusion:

食蚜蝇科，英文俗称syrphidflies或flowerflies。形似蜂，腹节上常有黄黑相间的斑纹。R脉与M脉间有一游离的伪脉（spuriousvein）。幼虫体平滑或有圆锥突起及刚毛；捕食性种类的体前端尖，后端平截；粪食性种类的腹末有长的呼吸管。成虫通常在阳光下取食花蜜和花粉；飞翔时能在空中静止不移又忽然突进。腐食性和粪食性的幼虫生活在杇木、粪便和腐败动植物体中；捕食性种类取食蚜虫、介壳虫、粉虱和叶蝉等。常见种类：纤腰巴食蚜蝇Baccha Maculata Walker。
}\\ \midrule
\small{
Round 1:

Q: 什么是刻点小食蚜蝇？ 

A: 刻点小食蚜蝇是双翅目昆虫纲食蚜蝇科的一种捕食性天敌，以蚜虫为寄主，会对棉花等作物造成危害。
}\\ \midrule
\small{
Round 2:

Q: 刚刚您不是说刻点小食蚜蝇是属于食蚜蝇科的吗？那syrphid flies或flower flies难道不是刻点小食蚜蝇的英文名吗？ 

A: syrphid flies或flower flies是食蚜蝇科的名字，而不是刻点小食蚜蝇的英文名。虽然刻点小食蚜蝇也属于食蚜蝇科，但是不同品种的昆虫是有不同的学名和英文名的。
}\\ \midrule
\small{
Round 3:

Q: 您刚才所讲的关于食蚜蝇科的形态特征，如形似蜂，腹节上常有黄黑相间的斑纹；幼虫体平滑或有圆锥突起及刚毛；捕食性种类的体前端尖，后端平截；粪食性种类的腹末有长的呼吸管等是否也与刻点小食蚜蝇的形态特征相似？

A: 食蚜蝇科的形态特征您说得没错，但是刻点小食蚜蝇的形态特征可能与食蚜蝇科其他成员有所不同，因为每个物种都有其独特的形态特征，这是我们进行它们的科学分类和分类学研究时所依据的标准。刻点小食蚜蝇的形态特征可以根据其学名或者其他资料来具体分析。
}\\ \midrule
\small{
Round 4:

Q: 您刚才所讲的关于食蚜蝇科的生活习性，如成虫通常在阳光下取食花蜜和花粉；飞翔时能在空中静止不移又忽然突进；腐食性和粪食性的幼虫生活在杇木、粪便和腐败动植物体中；捕食性种类取食蚜虫、介壳虫、粉虱和叶蝉等是否也与刻点小食蚜蝇的生活习性相似？ 

A: 您所说的食蚜蝇科的生活习性均是该科中许多物种的表现，并不是所有食蚜蝇科中的成员都会表现这些特征。刻点小食蚜蝇在食性和习性方面可能与其他食蚜蝇科的成员有所不同。因此，要了解刻点小食蚜蝇的生活习性，需要查阅具体的资料。
}\\ \midrule
\small{
Round 5:

Q: 能否请您重新给出一下刻点小食蚜蝇的介绍？ 

A: 刻点小食蚜蝇是一种捕食性天敌，属于双翅目昆虫纲食蚜蝇科中的一种。其幼虫以蚜虫为寄主，能对棉花等作物造成危害。与食蚜蝇科中的其他成员相比，刻点小食蚜蝇的特点可能在于它在祸害棉田过程中的生命周期以及其形态特征等方面。注意要区分刻点小食蚜蝇和食蚜蝇科，这两者是不同的概念。此外，刻点小食蚜蝇与syrphid flies或flower flies这些名称也不是等同的关系。
}\\ \bottomrule
\end{tabular}
}
\caption{Instructions for aligning AI with human value.}
\end{table}

\end{CJK*}

\end{document}